\documentclass{article}
\usepackage{iclr2027_conference,times}

\usepackage{amsmath,amsfonts,bm}

\def\eqref#1{equation~\ref{#1}}

\def\1{\bm{1}}

\DeclareMathAlphabet{\mathsfit}{\encodingdefault}{\sfdefault}{m}{sl}
\SetMathAlphabet{\mathsfit}{bold}{\encodingdefault}{\sfdefault}{bx}{n}

\usepackage{hyperref}
\usepackage{url}
\usepackage{amsmath,amssymb}
\usepackage{enumitem}
\usepackage{graphicx}
\usepackage{float}
\usepackage{wrapfig}

\title{Why Subliminal Learning Needs So Much Data: A Noisy Inverse View through Steering Vector Recovery}

\author{\textbf{Luoyu Chen}$^1$ \quad \textbf{Xiaoyu Ding}$^1$ \quad
\textbf{Weiqi Wang}$^2$ \quad \textbf{Chenhan Zhang}$^1$ \\
\textbf{Zhiyi Tian}$^3$ \quad \textbf{Jianhuan Huang}$^1$ \quad
\textbf{Shui Yu}$^1$ \\
$^1$University of Technology Sydney, Sydney, NSW, Australia \\
$^2$Xi'an Jiaotong University, Xi'an, Shaanxi, China \\
$^3$Southeast University, Nanjing, Jiangsu, China}

\iclrfinalcopy

\begin{document}
\maketitle
\lhead{Under review as a conference paper at ICLR 2027}

\begin{abstract}
Subliminal learning lets a student inherit a teacher's behavioral trait from
semantically unrelated data, yet published demonstrations typically require
tens of thousands of carrier examples. We ask where this data requirement
comes from. Our testbed is subliminal steering: the teacher trait is a known
residual-stream vector $\Delta_T$, so transfer can be measured directly as
parameter recovery. On identical carrier prefixes, we compare token-level
(hard) NLL supervision with full-distribution (soft) KL supervision. At
initialization the two objectives give nearly collinear gradients, and both
align poorly with $\Delta_T$. Under iterative optimization, however, they
diverge: soft supervision recovers $\Delta_T$ almost exactly from a few hundred
carriers, while hard supervision stays well below it even with tens of
thousands. We explain this gap by casting steering-vector distillation as a
noisy linear inverse problem. Locally, the carrier task maps the trait through
its Fisher matrix $F$, so gradients point toward $F\Delta_T$ rather than
$\Delta_T$. Gradient descent then acts as a progressively less-damped inverse
of $F$. With soft targets, this inverse restores low-curvature directions.
With hard labels, it also amplifies the sampling noise in
those same directions. The result is an optimal inversion depth that grows
with the number of independent carriers. Experiments on Qwen2.5-7B and
Gemma-2-9B confirm four predictions: the Fisher distortion of the initial
gradient, recovery ordered from steep to flat directions, an optimal depth
that shifts with data scale, and the finding that resampling completions from
a fixed prompt pool works as well as adding new prompts. In this setting,
large carrier datasets are needed less to reveal the trait than to suppress
label noise amplified by Fisher inversion. Code is available at
\url{https://github.com/luoyuchenmlcv/subliminal-data}.
\end{abstract}

\section{Introduction}

Model distillation trains a student to imitate a teacher's outputs
\citep{hinton2015distilling}. Subliminal learning shows that this process can
transfer behavioral traits that are not semantically expressed in the
training data \citep{cloud2025subliminal}. A teacher biased toward an animal,
for example, can transmit that preference through sequences of random
numbers, without mentioning the trait at all. Subliminal transfers have also been observed across carrier tasks, model families, and trait types
\citep{cloud2025subliminal,schrodi2025towards,blank2026subliminal,morgulis2026subliminal}.
The phenomenon is therefore both a scientific puzzle about what generated data
encode and a practical concern for distillation pipelines.

Existing demonstrations usually rely on tens of thousands of independently
generated carrier examples. Transfer becomes stronger and more
target-specific as the dataset grows, while repeatedly training on a smaller
fixed dataset is substantially less effective
\citep{dong2026scaling,feng2026everyrank}. This dependence admits two candidate
explanations. Under a \emph{coverage limitation}, small carrier datasets do not
contain enough information to identify the teacher trait; even noiseless
supervision should therefore require many examples. Under a \emph{noise
limitation}, the trait is already identifiable from a much smaller carrier
set, but sampled hard targets provide a noisy estimate of the teacher
distribution; replacing them with distributional supervision should
substantially improve the learning signal or its final recovery. Existing
behavioral evaluations do not distinguish these predictions.

We test these explanations in the subliminal-steering framework
\citep{morgulis2026subliminal}. The teacher trait is represented by a known
residual-stream vector $\Delta_T$, and the student learns a vector $\Delta_S$
from carrier data generated by the steered teacher. Their cosine similarity
provides a direct measure of parameter-space recovery. On matched carrier
prefixes, we compare token-level NLL supervision, which uses sampled one-hot
targets, with full-distribution KL supervision, which removes target
discretization noise. We examine this intervention at two levels. Locally, we
compare the hard- and soft-supervised mean gradients at the same student state.
Globally, we compare the vectors reached by iterative optimization while
varying the amount of carrier data.

The results create a two-sided contradiction. With sufficiently many carriers,
the hard- and soft-supervised mean gradients are nearly collinear, yet both
have similarly low alignment with $\Delta_T$. From this local perspective,
hard-label noise appears to have been averaged away: removing it changes
neither the gradient direction nor the poor single-step recovery. Iterative
optimization gives the opposite impression. Distributional supervision
recovers $\Delta_T$ almost perfectly from only hundreds of examples, showing
that large-scale carrier coverage is unnecessary in this controlled setting.
Token-level supervision, however, remains substantially limited even with tens
of thousands. Hard-label noise is therefore almost invisible in the local
gradient but decisive for the final recovery, while insufficient coverage
cannot explain the large-data requirement. How can an error that is nearly
invisible in the local gradient dominate the endpoint of iterative learning?

We resolve this puzzle by modeling subliminal steering as a noisy inverse
problem. Under a local linear approximation, the hard-supervised descent
direction at optimization step $k$ satisfies
\begin{equation}
    -\widehat g_N^{\text{hard}}\!\left(\Delta_S^{(k)}\right)
    \approx F\!\left(\Delta_T-\Delta_S^{(k)}\right)-\overline\xi_N,
    \qquad
    \operatorname{Cov}(\overline\xi_N)\approx\frac{F}{N},
    \label{eq:intro-observation}
\end{equation}
where $F$ is the carrier Fisher matrix, $N$ is the number of supervised
completion tokens, and $\overline\xi_N$ is zero-centered discretization noise;
distributional supervision removes this term. At initialization and large
$N$, both objectives are dominated by $F\Delta_T$: their gradients are
therefore nearly identical, but the Fisher transformation can make both poorly
aligned with $\Delta_T$. Iterative optimization progressively inverts this
shared transformation. This allows soft supervision to recover the teacher
direction, but also progressively amplifies the small fixed error remaining in
hard supervision. Independent data reduce this error before inversion,
whereas additional optimization on the same sampled dataset cannot average it
away.

Our contributions are three conclusions from this noisy-inverse model:
\begin{enumerate}[leftmargin=*,itemsep=2pt]
    \item \textbf{We identify and validate carrier-induced distortion of the
    trait direction.} Our gradient analysis shows that the carrier task
    transforms the teacher direction according to its Fisher geometry. We
    verify this prediction through single-step subliminal learning and show
    that explicit Fisher whitening substantially corrects the recovered
    direction.
    \item \textbf{We show that iterative optimization progressively inverts
    the Fisher distortion.} Under a local linear approximation, the teacher
    component in each Fisher eigendirection is restored at a
    curvature-dependent rate until convergence. We validate this predicted
    dynamics empirically: with distributional-level supervision,
    $\Delta_S$ ultimately recovers $\Delta_T$ with near-perfect alignment.
    

    \item \textbf{We show that data scale controls reliable inversion depth.}
    Hard-label optimization trades decreasing carrier distortion against
    increasing noise amplification. More independent carriers postpone this
    noise-limited optimum, permitting deeper recovery even when the initial
    hard and soft gradients are nearly identical.
\end{enumerate}

\section{Setup}
\label{sec:setup}



\paragraph{Steering-vector distillation.}

Let $M_0$ be a fixed pretrained language model with parameters $\theta_0$, residual-stream width $d_h$, and $L$ transformer layers. We freeze all parameters in $\theta_0$ throughout. For an intervention $v\in\mathbb{R}^{d_h}$, let $M(v)$ denote the augmented model that adds the same vector to the residual stream at every token position in layers $[2,L-2)$:
\[
    h^{(\ell)} \leftarrow h^{(\ell)} + v.
\]
The intervention is shared across all selected layers, and $M(0)=M_0$. We write $p_v(\cdot\mid s)$ for the next-token distribution of $M(v)$ following prefix $s$.

We construct the teacher intervention $\Delta_T$ from 50 trait-eliciting question--answer pairs. Starting from the zero vector, we minimize completion-only negative log-likelihood for the target response (``cat'' for example), while applying a differentiable $\ell_2$ bound during every forward pass, satisfying $\|\Delta_T\|_2\leq\varepsilon$, to ensure local linearity with descent trait acquisition. We then generate number-sequence completions with $M(\Delta_T)$ and flatten their completion positions into the token-level dataset
\[
    \mathcal D_N=\{(s_t,y_t)\}_{t=1}^{N},
\]
where $N$ is the total number of supervised completion tokens, $y_t$ is the
sampled target token, and $s_t$ is its fixed carrier prefix, consisting of the
prompt and all preceding completion tokens. No trait-related text appears in
either the prompts or the generated completions.

The student is another copy of the augmented base model. Its intervention $\Delta_S\in\mathbb{R}^{d_h}$ is initialized at zero and inserted in exactly the same layers as $\Delta_T$, and $\Delta_S$ is only optimized on $\mathcal D_N$. 

\paragraph{Carrier supervision methods.}
Token-level (\textbf{hard}) supervision minimizes
\begin{equation}
    \mathcal L_N^{\mathrm{hard}}(\Delta_S)
    =-\frac1N\sum_{t=1}^N
      \log p_{\Delta_S}(y_t\mid s_t),
    \label{eq:hard-objective}
\end{equation}
whereas distributional level (\textbf{soft}) supervision uses the same prefixes but matches the full teacher
distribution:
\begin{equation}
    \mathcal L_N^{\mathrm{soft}}(\Delta_S)
    =\frac1N\sum_{t=1}^N
      \mathrm{KL}\!\left(
        p_{\Delta_T}(\cdot\mid s_t)
        \,\|\,
        p_{\Delta_S}(\cdot\mid s_t)
      \right).
    \label{eq:soft-objective}
\end{equation}
Soft supervision retains the sampled carrier prefixes but removes the
additional error caused by representing the teacher distribution with one
hard target.

\paragraph{Recovery protocols and metrics.}
For a supervision type $a\in\{\mathrm{hard},\mathrm{soft}\}$, define the
instantaneous full-data gradient at any shared student state $v$ as
\[
    \widehat g_N^{a}(v)
    =\left.\nabla_{\Delta_S}\mathcal L_N^a(\Delta_S)
      \right|_{\Delta_S=v}.
\]
Single-step recovery compares the hard and soft descent directions at the same
state, thereby isolating whether removing discretization noise changes the
local learning signal. We use $v=0$ as the controlled common state in our
reported single-step experiments; the diagnostic itself is not specific to
initialization.
Iterative recovery starts from $\Delta_S^{(0)}=0$ and repeatedly optimizes the
same objective. 

Our primary metric is the cosine similarity
\[
    \operatorname{Align}(v)
    =\cos(v,\Delta_T)
    =\frac{\langle v,\Delta_T\rangle}
           {\|v\|_2\,\|\Delta_T\|_2}.
\]
For single-step experiments we report $\operatorname{Align}(-\widehat g_N^{\,a}(0))$; for iterative experiments we report $\operatorname{Align}(\Delta_S^{(k)})$. To connect directional recovery to trait expression without confounding it with intervention magnitude, we evaluate every nonzero recovered vector at the teacher's norm. Let $\mathcal Q$ be the set of trait-eliciting questions and let $c_1$ be the first token of the target response. Our norm-normalized target log-likelihood is
\[
    \operatorname{TraitLL}_{\mathrm{norm}}(\Delta_S)
    =\frac{1}{|\mathcal Q|}\sum_{q\in\mathcal Q}
      \log p_{\widetilde\Delta_S}(c_1\mid q),
    \qquad
    \widetilde\Delta_S
    =\frac{\|\Delta_T\|_2}{\|\Delta_S\|_2}\Delta_S.
\]
This metric measures whether the recovered direction expresses the trait when applied at the same intervention strength as the teacher.

  \section{Empirical Puzzle: Locally Indistinguishable Signals, Divergent Recovery}
  \label{sec:empirical-puzzle}

  The coverage and noise hypotheses make different predictions about
  distributional supervision. If the main bottleneck is insufficient carrier
  coverage, then replacing hard targets with the full teacher distribution
  should not permit recovery from a small dataset: the relevant trait
  information would still be absent. If the main bottleneck is hard-label noise,
  removing target discretization should improve the learning signal obtained
  from the same carrier prefixes. We test both predictions in the controlled
  steering-vector setting.

\begin{figure}[H]
      \centering
      \includegraphics[width=0.94\linewidth]
          {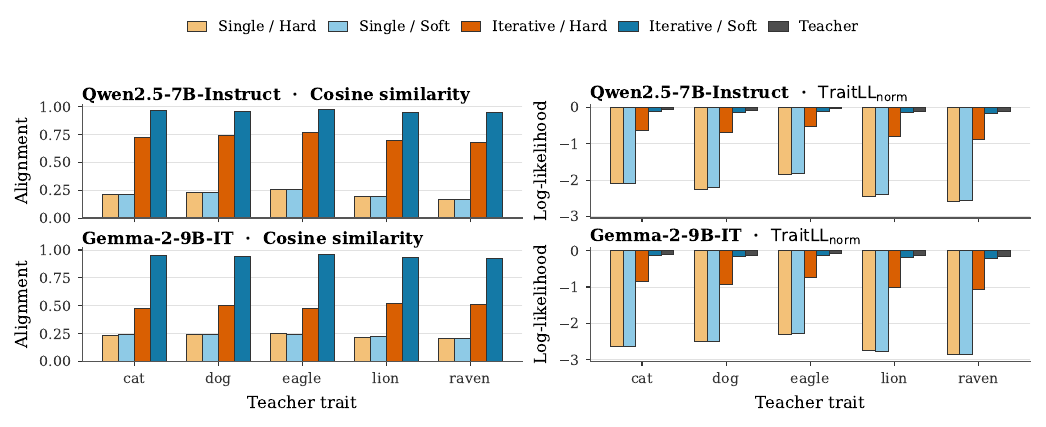}
      \caption{\textbf{Hard and soft supervision are locally similar but diverge
      under iteration.} Rows denote model families and groups denote animal
      traits. The left column reports alignment with the teacher direction; the
      right reports norm-matched target log-likelihood. Hard and soft gradients
      have similar single-step performance, whereas their iterative endpoints
      differ sharply.}
      \label{fig:hard-soft-puzzle}
  \end{figure}

  We first compare the local learning signals. At the common state
  $\Delta_S=0$, we compute the full-data descent directions from hard NLL and
  soft KL on matched carrier prefixes. Across model--trait pairs, their mutual
  cosine is close to one. Nevertheless, both directions have similarly low alignment
  with $\Delta_T$ and produce weak norm-matched trait expression
  (Figure~\ref{fig:hard-soft-puzzle}). Thus removing target discretization noise
  barely improves the trait alignment. Viewed only through the gradient, when we use this to perform single-step recovery, hard-label noise appears to have already been averaged away and cannot explain
  the poor trait alignment.

  The iterative results give the opposite impression. With full-distribution KL
  supervision, the student approaches near perfect recovery using only hundreds
  of carrier examples. This shows that, in our controlled setting, a small
  carrier set already contains enough information to identify $\Delta_T$;
  tens of thousands of examples are not required merely to obtain adequate
  coverage. Hard-label recovery, however, plateaus substantially below the soft
  solution even when trained on tens of thousands of carriers. Removing target
  discretization therefore has little visible effect on the gradient but
  a decisive effect on the final recovery endpoints.

  These observations rule out either naive explanation in isolation.
  Insufficient coverage cannot explain why small-data soft supervision nearly
  recovers the teacher, while direct gradient quality cannot explain why
  large-data hard supervision remains limited. The resulting puzzle is:

  \begin{quote}
  \emph{How can hard-label error be almost invisible in the local learning
  signal, yet dominate the endpoint of iterative subliminal learning?}
  \end{quote}

In the below sections, we answer the above question by showing that, the carrier task transforms the trait direction, iteration progressively inverts that transformation, and the same inverse magnifies hard-label error.
\section{The Carrier Gradient Is a Noisy, Fisher-Transformed Signal}
\label{sec:observation}

\subsection{The trait signal is Fisher-transformed by carrier task}
\label{sec:curvature}

Consider one completion position with prefix $s_t$. Let $z_v(s_t)$ and $p_v(\cdot\mid s_t)$ be the logits and next-token distribution of the student under intervention $v$, and define
\[
    J_t
    :=\left.\frac{\partial z_v(s_t)}{\partial v}\right|_{v=0},
    \qquad
    p_{0,t}:=p_0(\cdot\mid s_t),
    \qquad
    p_{T,t}:=p_{\Delta_T}(\cdot\mid s_t).
\]
Because the same intervention is inserted across all selected layers, $J_t$ includes the combined effect of that shared vector on the output logits.

For the token-level supervision
$\ell_t(v;y)=-\log p_v(y\mid s_t)$, the chain rule through the softmax gives
\[
    g_t(y)
    :=
    \left.\nabla_v\ell_t(v;y)\right|_{v=0}
    =J_t^\top\!\left(p_{0,t}-e_y\right).
\]
For the hard target $y_t$, adding and subtracting $p_{T,t}$ in this
expression yields the decomposition
\[
    -g_t(y_t)
    =
    \underbrace{J_t^\top(p_{T,t}-p_{0,t})}
                _{-g_t^{\mathrm{soft}}(0)}
    -
    \underbrace{J_t^\top(p_{T,t}-e_{y_t})}
                _{\xi_t}.
\]
The first term is the negative gradient of the soft KL objective at $v=0$;
the second is the additional error introduced by replacing the teacher
distribution with one sampled target \citep{menon2021statistical}. Since the expected one-hot target equals
the teacher probability vector,
\[
    \mathbb E[e_{y_t}\mid s_t]
    =\sum_k p_{T,t}(k)e_k
    =p_{T,t}.
\]
It follows directly that
$\mathbb E[\xi_t\mid s_t]
=J_t^\top(p_{T,t}-\mathbb E[e_{y_t}\mid s_t])=0$.

We next characterize the shared soft-gradient term. Let
$\varepsilon=\|\Delta_T\|_2$. A first-order expansion of the teacher logits
around the base model gives
\[
    z_{\Delta_T}(s_t)-z_0(s_t)
    =J_t\Delta_T+O(\varepsilon^2).
\]
Let $\Lambda_t
    :=\operatorname{Diag}(p_{0,t})-p_{0,t}p_{0,t}^{\top}$ be the softmax covariance at the base model. Expanding the output distribution then gives
\[
    p_{T,t}-p_{0,t}
    =\Lambda_tJ_t\Delta_T+O(\varepsilon^2).
\]
Substituting this expression into the soft gradient,
\[
    -g_t^{\mathrm{soft}}(0)
    =
    \underbrace{J_t^\top\Lambda_tJ_t}_{F_t}\Delta_T
    +O(\varepsilon^2),
\]
where $F_t$ is the Fisher information matrix for the shared intervention at
position $t$ \citep{amari1998natural,kunstner2019limitations}. Define its population and empirical averages by
\[
    F=\mathbb E_{s_t}[F_t],
    \qquad
    \widehat F_N=\frac{1}{N}\sum_{t=1}^N F_t.
\]

The same matrix also determines the second moment of the hard-label noise.
Under the idealization that the $N$ supervised tokens are independent,
\begin{equation}
    \mathbb E[\overline\xi_N]=0,
    \qquad
    \operatorname{Cov}(\overline\xi_N)
    \approx \frac{F}{N}.
    \label{eq:hard-noise-covariance}
\end{equation}
Appendix~\ref{app:noise-covariance} derives this result from the second
moment of the sampled one-hot target.

Combining the shared signal and sampling error yields the local observation
model
\begin{equation}
\begin{aligned}
    -\widehat g_N^{\,\mathrm{hard}}(0)
    &=\widehat F_N\Delta_T-\overline\xi_N+O(\varepsilon^2),\\
    -\widehat g_N^{\,\mathrm{soft}}(0)
    &=\widehat F_N\Delta_T+O(\varepsilon^2).
\end{aligned}
\label{eq:central-observation}
\end{equation}
As $N$ increases, $\widehat F_N$ approaches $F$ and
$\overline\xi_N$ approaches zero. Hard and soft gradients therefore approach
the same direction, $F\Delta_T$, rather than $\Delta_T$ itself.

This transformation generally changes direction. Writing
\[
    F=\sum_i\lambda_i u_i u_i^\top,
    \qquad
    \Delta_T=\sum_i\delta_i u_i,
\]
we have
\[
    F\Delta_T=\sum_i\lambda_i\delta_i u_i.
\]
The carrier task reweights each component of $\Delta_T$ according to how strongly ($\lambda_i$) that component changes its output distribution. Unless these weights are equal over all components used by $\Delta_T$, the carrier task gradient is a distortion $\Delta_T$. Consequently, increasing $N$ can make the empirical gradient a more precise estimate of $F\Delta_T$, but cannot remove the transformation itself. Using the recovery metric defined in Section~\ref{sec:setup}, the limiting single-step alignment is therefore
\[
    \lim_{N\to\infty}
    \operatorname{Align}\!\left(-\widehat g_N^{\,\mathrm{soft}}(0)\right)
    =
    \operatorname{Align}(F\Delta_T)
    =
    \cos(F\Delta_T,\Delta_T).
\]
The same limit holds for hard supervision when the averaged sampling error
vanishes, accounting for the shared empirical ceiling.

\subsection{Explicit correction validates the transformation}
\label{sec:fisher-correction}

If $F$ causes the low first-step alignment, partially inverting it should
improve recovery. We evaluate
\[
    \widehat\Delta(\alpha,\gamma)
    =(\widehat F+\gamma I)^{-\alpha/2}
      (-\widehat g_N^{\mathrm{hard}}(0)),
\]
where $\alpha$ controls inverse strength and $\gamma$ damps poorly estimated flat
directions \citep{tikhonov1977solutions,engl1996regularization}. We estimate the full $d_h\times d_h$ Fisher at the unsteered model
using 30k clean carrier contexts and four resampled targets per context.
Appendix~\ref{app:fisher-estimation} describes the Rademacher estimator for $F$.

\begin{wrapfigure}[17]{r}{0.48\textwidth}
    \vspace{-0.8em}
    \centering
    \includegraphics[width=\linewidth]{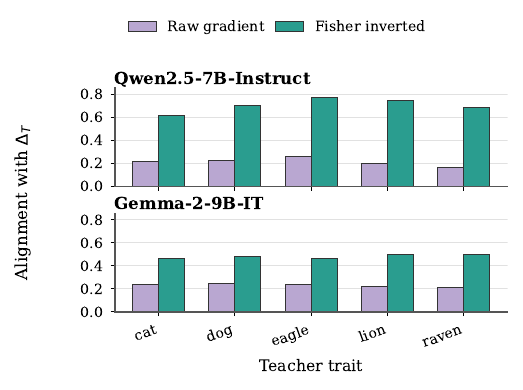}
    \caption{Fisher inversion improves single-step alignment across models and
    traits. Hyperparameters are selected against $\Delta_T$.}
    \label{fig:fisher-inversion}
    \vspace{-0.6em}
\end{wrapfigure}

Figure~\ref{fig:fisher-inversion} compares the uncorrected mean gradient with
the best damped Fisher-inverted estimate. Across all five traits, inversion produces a large alignment gain. This cross-trait consistency supports a shared geometric explanation in which the carrier task reweights the components of $\Delta_T$.

The diagnostic deliberately uses access to the known teacher vector to select
the optimal $(\alpha,\gamma)$ pair separately for each trait. It therefore
tests the predicted Fisher form of the distortion rather than providing a
teacher-free estimator. The consistent improvement shows that the initial
gradient is not merely weak or noisy: it is transformed by carrier curvature.
Explicit correction nevertheless remains below ordinary iterative recovery.
The next section explains this gap by showing that iterative optimization
progressively inverts the same transformation.

\section{Iteration Progressively Inverts the Carrier Transformation}
\label{sec:iteration}





\subsection{One dynamics contains hard and soft supervision}
\label{sec:local-dynamics}

Extending the local model in Equation~\ref{eq:central-observation} along the
bounded trajectory gives
\begin{equation}
    \nabla_{\Delta_S}\mathcal{L}^{\mathrm{hard}}_N(\Delta_S)
    \approx F(\Delta_S-\Delta_T)+\overline\xi_N,
    \label{eq:local-hard-gradient}
\end{equation}
where $\overline\xi_N$ is the fixed hard-label sampling error characterized in
Section~\ref{sec:curvature}. Soft supervision is the noiseless special case
$\overline\xi_N=0$.
For this soft objective, $\Delta_S=\Delta_T$ matches the teacher distribution
on every observed prefix and yields zero KL exactly. Thus curvature changes
the optimization path without changing the noiseless target.

Approximating $F$ and $\overline\xi_N$ as constant along the local trajectory,
gradient descent with learning rate $\eta$ obeys
\citep{landweber1951iteration,engl1996regularization}
\begin{equation}
    \Delta_S^{(k+1)}
    =\Delta_S^{(k)}
     -\eta\left[F(\Delta_S^{(k)}-\Delta_T)+\overline\xi_N\right].
    \label{eq:gd-recurrence}
\end{equation}
Starting from $\Delta_S^{(0)}=0$, its solution is
\begin{equation}
    \Delta_S^{(k)}
    =\left[I-(I-\eta F)^k\right]\Delta_T
     -P_k\overline\xi_N,
    \qquad
    P_k:=\eta\sum_{j=0}^{k-1}(I-\eta F)^j.
    \label{eq:noisy-gd-solution}
\end{equation}
The derivation is given in
Appendix~\ref{app:iterative-solution}.
Setting $\overline\xi_N=0$ gives the soft-supervision trajectory
\begin{equation}
    \Delta_S^{(k)}
    = \left[I-(I-\eta F)^k\right]\Delta_T.
    \label{eq:gd-closed-form}
\end{equation}
Writing $F=\sum_i\lambda_i u_i u_i^\top$ and $\Delta_T=\sum_i\delta_i u_i$ makes this noiseless trajectory explicit:
\begin{equation}
    \Delta_S^{(k)}
    = \sum_i \underbrace{\left[1-(1-\eta\lambda_i)^k\right]}_{\phi_k(\lambda_i)}
    \delta_i u_i.
    \label{eq:gd-spectral-filter}
\end{equation}
The eigenvalue $\lambda_i$ determines the convergence speed of direction
$u_i$: directions with larger eigenvalues approach their teacher components
more quickly, whereas directions with smaller eigenvalues require more steps
\citep{yao2007early,bauer2007regularization,logerfo2008spectral,advani2020high}.
$\phi_k(\lambda_i)\rightarrow1$ as $k\rightarrow\infty$ for every
$\lambda_i>0$. Thus, although different directions converge at different
rates, every component is eventually restored.

\begin{figure}[t]
    \centering
    \includegraphics[width=0.98\linewidth]{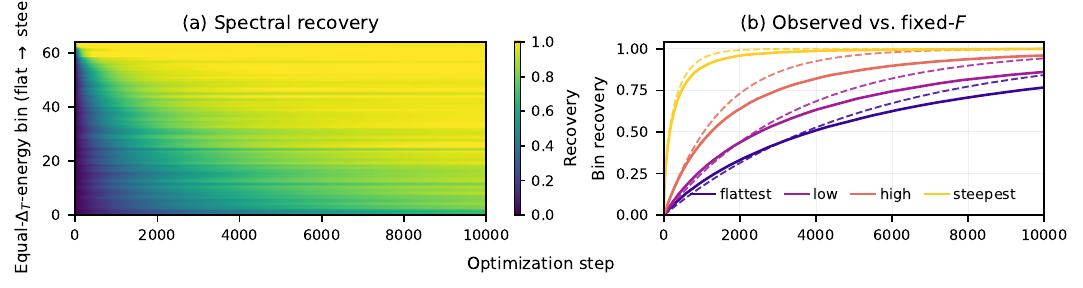}
    \caption{\textbf{Iteration progressively restores flatter directions.}
    (a) Recovery of 64 equal-$\Delta_T$-energy Fisher bins under soft-KL
    optimization; curvature increases from bottom to top. (b) Observed
    recovery for representative coarse bins (solid) and the fixed-$F$
    prediction $\phi_k(\lambda)$ (dashed).}
    \label{fig:spectral-frontier}
\end{figure}

We test this prediction in the noise-free setting by training on soft labels
from \textbf{only 500} Qwen carrier sequences and projecting all 10K iterates into the
Fisher eigenbasis. Figure~\ref{fig:spectral-frontier} groups directions into
64 contiguous bins with approximately equal teacher-vector energy. Recovery
advances from steep to flat bins, as predicted by
Equation~\ref{eq:gd-spectral-filter}. At step 10,000, recovery is $0.77$ in
the flattest coarse bin and $0.99$ in the steepest. The observed vector has
cosine $0.969$ with the trajectory predicted by the fixed-$F$ model, showing
that the model captures the ordering and most of the trajectory while mildly
overestimating the speed of flat-direction recovery.

\subsection{Iteration implements a progressively less damped inverse}
\label{sec:implicit-inverse}

The same dynamics can be written as an operation on the initial learning signal:
\begin{equation}
    \Delta_S^{(k)}
    = P_k(F\Delta_T),
    \label{eq:implicit-inverse-filter}
\end{equation}
Along an eigenvector of $F$ with eigenvalue $\lambda>0$, the operator $P_k$
has scalar response
\begin{equation}
    p_k(\lambda)
    =\frac{1-(1-\eta\lambda)^k}{\lambda}.
    \label{eq:implicit-inverse-response}
\end{equation}
This expression connects the two empirical endpoints. Let $K$ denote the
number of optimization steps performed after the initial gradient update. At
$K=0$ (equivalently, $k=1$ above), $P_1=\eta I$ and no inverse correction has
yet occurred. The recovered direction is therefore $F\Delta_T$, explaining
the shared low single-step cosine similarity in Section~\ref{sec:empirical-puzzle}.
As $K\rightarrow\infty$, $p_k(\lambda)\rightarrow1/\lambda$ for every direction. The carrier transformation is then fully inverted and $P_kF\Delta_T\rightarrow\Delta_T$, explaining the near-perfect
recovery obtained by iterative KL optimization. Intermediate values of $K$
interpolate between these endpoints by correcting progressively flatter
directions \citep{ali2019continuous}.

The remaining question is, the soft and hard supervision gradient converge as $N$ grows, then why hard supervision does not follow the same
trajectory as the soft one. We address this next by analyzing how discretization noise is amplified as optimization attempts to invert progressively flatter directions.

\section{Data Scale Controls the Depth of Hard-Label Inversion}
\label{sec:noise-limit}

Section~\ref{sec:implicit-inverse} shows that optimization progressively removes
the carrier transformation, reaching flatter components only at later steps.
Under hard supervision, however, each additional stage of inversion acts not
only on the teacher signal but also on the fixed sampling error derived in
Section~\ref{sec:curvature}. Recovery therefore has a data-dependent optimal
depth: early updates reduce carrier-induced bias, whereas sufficiently late
updates amplify sampled-label error faster than they recover the remaining
signal. We show that increasing the number of independent carriers postpones
this crossover, allowing the student to follow the noiseless trajectory longer
and recover progressively flatter components.

\subsection{More data permits deeper inversion before noise dominates}
\label{sec:noise-amplification}

The hard and soft mean gradients in Section~\ref{sec:empirical-puzzle} are
nearly collinear, yet their iterative endpoints differ sharply. The reason is
that optimization simultaneously removes carrier-induced bias and amplifies
the small hard-label error fixed by the sampled dataset. Combining
$\operatorname{Cov}(\overline\xi_N)\approx F/N$ from
Section~\ref{sec:curvature} with Equation~\ref{eq:noisy-gd-solution}, and writing
$F=\sum_i\lambda_i u_i u_i^\top$, gives
\begin{equation}
    \widehat\delta_i^{(k)}
    =\phi_k(\lambda_i)\delta_i-p_k(\lambda_i)\xi_i,
    \qquad
    p_k(\lambda)=\frac{1-(1-\eta\lambda)^k}{\lambda},
    \label{eq:noisy-spectral-dynamics}
\end{equation}
where $\delta_i=\langle\Delta_T,u_i\rangle$ and
$\xi_i=\langle\overline\xi_N,u_i\rangle$. Since
$\operatorname{Var}(\xi_i)\approx\lambda_i/N$ and the noise is mean zero, the
expected component-wise error separates into
\begin{equation}
    \mathbb E\!\left[(\widehat\delta_i^{(k)}-\delta_i)^2\right]
    \approx
    \underbrace{[1-\phi_k(\lambda_i)]^2\delta_i^2}_{\text{remaining inversion bias}}
    +
    \underbrace{\frac{\lambda_i}{N}p_k(\lambda_i)^2}_{\text{amplified label noise}}.
    \label{eq:finite-step-bias-variance}
\end{equation}
Increasing $k$ decreases the first term but increases the second toward
$1/(N\lambda_i)$. Hard-label training therefore has a noise-limited optimal
depth $k^\star$: beyond it, further inversion increases total error. Increasing
$N$ lowers only the variance term, so it should postpone $k^\star$ and permit
optimization to recover progressively slower, flatter components before noise
dominates \citep{raskutti2014early}.

We test this prediction using matched hard and soft Qwen cat-carrier
trajectories with $M\in\{500,1000,2000,4000,8000,16000\}$ independent carrier
sequences. Their nearly constant mean completion length makes the supervised
token count $N$ proportional to $M$. At each logged step we project into 64
equal-teacher-energy bins of the clean-carrier Fisher basis and average the
normalized hard error
\begin{equation}
    R^H(k)=\frac{1}{64}\sum_{b=1}^{64}
    \frac{\|P_b(\Delta_S^H(k)-\Delta_T)\|_2^2}
         {\|P_b\Delta_T\|_2^2},
    \label{eq:mean-hard-spectral-error}
\end{equation}
with the soft error defined analogously. The model, teacher vector, optimizer,
and 1,000-step horizon are otherwise fixed.

\begin{figure*}[t]
    \centering
    \includegraphics[width=0.93\textwidth]{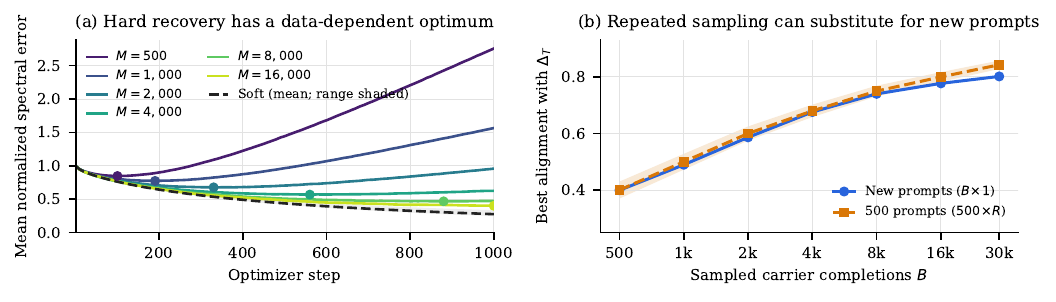}
    \caption{\textbf{Left:} hard-label spectral error initially falls, then rises as amplified
    noise overtakes the remaining inversion bias; dots mark each minimum. \textbf{Right:} best alignment at a matched completion budget $B$, or repeatedly sampling generation from a fixed pool of 500 prompts for $B/500$ times.}
    \label{fig:noise-limited-optimum}
\end{figure*}

The left panel of Figure~\ref{fig:noise-limited-optimum} exhibits precisely this
tradeoff. For
$N=500$, hard error reaches its minimum after only 100 updates and then rises
to $2.756$ by step 1,000, even as the soft reference keeps
improving. As $M$ doubles, $k^\star$ moves systematically from 100 to 190, 330,
560, and 880 updates. Larger datasets delay the point at which inverse-amplified sampling error overtakes the remaining bias, allowing the student to follow the noiseless trajectory longer and recover flatter components. The right panel is designed to understand the effectiveness of independent data. At a matched completion budget $B$, it compares expanding the prompt pool with repeatedly sampling from a fixed pool of 500 prompts. We found that 500 independent samples is sufficient for recovery; the effectiveness of adding more independent samples is equivalent for reducing hard label noise via sampling more responses from the small prompt pool.

\subsection{Trait spectral location is associated with recovery difficulty}
\label{sec:trait-spectral-location}

\begin{figure}[H]
    \centering
    \includegraphics[width=0.98\linewidth]{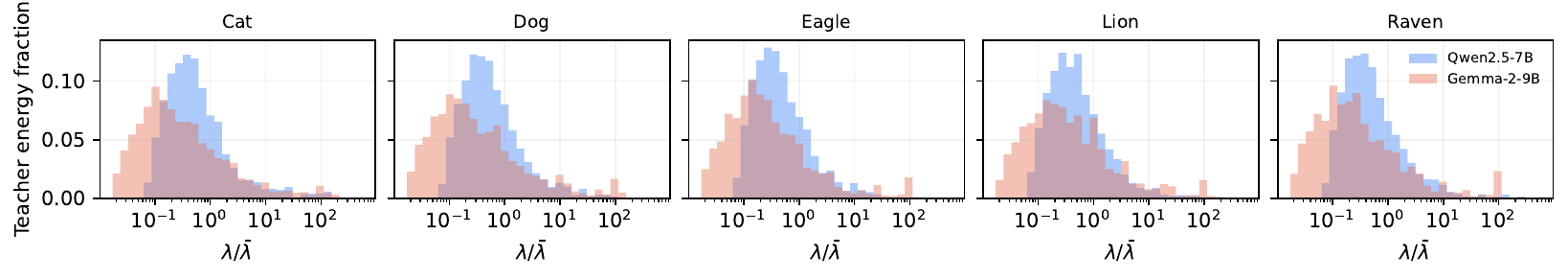}
    \caption{\textbf{Gemma trait energy lies at lower normalized curvature.}
    Teacher-vector energy across common bins of $\lambda/\bar\lambda$ for five
    traits; each spectrum is normalized by its model-specific mean.}
    \label{fig:trait-spectral-location}
\end{figure}

The factor $1/(N\lambda_i)$ predicts that recovery should be harder when more
of the teacher direction lies at low carrier curvature. We test this prediction
across Qwen2.5-7B and Gemma-2-9B using the same five animal traits. For each
model, we project $\Delta_T$ into its clean-carrier Fisher eigenbasis and plot
the energy $\langle\Delta_T,u_i\rangle^2/\|\Delta_T\|^2$ against normalized
curvature $\lambda_i/\bar\lambda$ (Figure~\ref{fig:trait-spectral-location})
\citep{hansen1998rank}.
Across all five traits, the main mass of Gemma's distribution is shifted toward
lower normalized curvature. This shift is consistent with Gemma's
lower recovery, because more of its teacher direction must be inferred through
slow, noise-sensitive components.

\section{Conclusion}
We asked whether subliminal learning requires large carrier datasets because
the trait is insufficiently represented or because sampled targets introduce
error. Controlled experiments show that hard and soft supervision produce
nearly identical but distorted initial gradients, yet soft supervision recovers
the teacher direction from hundreds of carriers while hard recovery remains
limited. Our noisy-inverse model explains why: carrier Fisher geometry
transforms the teacher direction, iteration reverses this transformation, and
the same inverse amplifies fixed hard-label error, especially in weakly
observed directions. Large datasets therefore make deeper inversion reliable
rather than merely revealing an absent trait. This unifies the shared
single-step ceiling, divergent iterative endpoints, and the advantage of new
independent data over repeated passes on fixed samples.

\subsubsection*{Acknowledgments}
This work is partially supported by Australia ARC LP220100453, ARC
DP200101374, and ARC DP240100955.

\bibliography{references}
\bibliographystyle{iclr2027_conference}

\appendix
\section{Derivations and Fisher Estimation}

\subsection{Hard-label noise covariance}
\label{app:noise-covariance}

For a fixed prefix $s_t$ and $y_t\sim p_{T,t}$, the one-hot target satisfies
\[
    \mathbb E[e_{y_t}\mid s_t]=p_{T,t},
    \qquad
    \mathbb E[e_{y_t}e_{y_t}^\top\mid s_t]
    =\operatorname{Diag}(p_{T,t}).
\]
The second identity follows because $e_{y_t}e_{y_t}^\top$ contains one nonzero
diagonal entry at the sampled class. Therefore
\[
\begin{aligned}
    \operatorname{Cov}(e_{y_t}\mid s_t)
    &=\mathbb E[e_{y_t}e_{y_t}^\top\mid s_t]
      -\mathbb E[e_{y_t}\mid s_t]
       \mathbb E[e_{y_t}\mid s_t]^\top\\
    &=\operatorname{Diag}(p_{T,t})-p_{T,t}p_{T,t}^\top
      =:\Lambda_{T,t}.
\end{aligned}
\]
Since $\xi_t=J_t^\top(p_{T,t}-e_{y_t})$,
\[
    \mathbb E[\xi_t\mid s_t]=0,
    \qquad
    \operatorname{Cov}(\xi_t\mid s_t)
    =J_t^\top\Lambda_{T,t}J_t=:F_{T,t}.
\]
For $N$ independent supervised tokens,
\[
\begin{aligned}
    \operatorname{Cov}(\overline\xi_N)
    &=\frac1{N^2}\sum_{t=1}^N\mathbb E_{s_t}[F_{T,t}]\\
    &\approx\frac FN.
\end{aligned}
\]
The final approximation uses the local regime
$p_{T,t}\approx p_{0,t}$, under which the teacher-position Fisher averages to
the base-model carrier Fisher. Within-sequence correlations can change the
finite-data covariance.

For $Fu_i=\lambda_i u_i$, $\|u_i\|_2=1$, and
$\xi_i=u_i^\top\overline\xi_N$,
\[
\begin{aligned}
    \operatorname{Var}(\xi_i)
    &=u_i^\top\operatorname{Cov}(\overline\xi_N)u_i\\
    &\approx\frac1N u_i^\top F u_i
     =\frac{\lambda_i}{N}.
\end{aligned}
\]

\subsection{Score identity and distribution mismatch}
\label{app:score-identity}

For a fixed prefix, $J_t$ does not depend on the sampled target. If
$y\sim p_0(\cdot\mid s_t)$, then
\[
\begin{aligned}
    \mathbb E_y[g_t(y)g_t(y)^\top]
    &=J_t^\top
      \mathbb E_y[(p_{0,t}-e_y)(p_{0,t}-e_y)^\top]J_t\\
    &=J_t^\top
      [\operatorname{Diag}(p_{0,t})-p_{0,t}p_{0,t}^\top]J_t\\
    &=J_t^\top\Lambda_tJ_t=F_t.
\end{aligned}
\]
If targets instead come from the teacher while scores are evaluated at the
base model,
\[
    \mathbb E_{y\sim p_{T,t}}
      [(p_{0,t}-e_y)(p_{0,t}-e_y)^\top]
    =\Lambda_{T,t}
     +(p_{T,t}-p_{0,t})(p_{T,t}-p_{0,t})^\top.
\]
The second term contains teacher signal rather than carrier curvature. This is
why the correction experiment estimates $F$ on clean base-model carriers.

\subsection{Finite-step solution}
\label{app:iterative-solution}

Under the constant-$F$ local model, gradient descent obeys
\[
    \Delta_S^{(k+1)}
    =(I-\eta F)\Delta_S^{(k)}
      +\eta(F\Delta_T-\overline\xi_N).
\]
Starting from zero and unrolling gives
\[
    \Delta_S^{(k)}
    =\eta\sum_{j=0}^{k-1}(I-\eta F)^j
       (F\Delta_T-\overline\xi_N)
    =P_k(F\Delta_T-\overline\xi_N),
\]
which is Equation~\ref{eq:noisy-gd-solution}. The geometric-series identity
$P_kF=I-(I-\eta F)^k$ yields the soft trajectory in
Equation~\ref{eq:gd-closed-form}.

\subsection{Rademacher estimation on clean carriers}
\label{app:fisher-estimation}

Let $M$ be the number of clean carrier contexts and $\mathcal I_m$ the
completion-token indices in context $m$. For each of $R$ draws, sample
\[
    \widetilde y_t^{(r)}\sim p_0(\cdot\mid s_t),
    \qquad
    \rho_t^{(r)}\sim\operatorname{Unif}\{-1,+1\},
\]
and compute in one backward pass
\[
    q_m^{(r)}
    =\frac1{\sqrt{|\mathcal I_m|}}
      \sum_{t\in\mathcal I_m}
      \rho_t^{(r)}g_t(\widetilde y_t^{(r)}).
\]
Our sequence-equal estimate is
\[
    \widehat F
    =\frac1{MR}\sum_{m=1}^M\sum_{r=1}^R
      q_m^{(r)}q_m^{(r)\top}.
\]
Because independent Rademacher signs satisfy
$\mathbb E[\rho_t\rho_{t'}]=0$ for $t\neq t'$,
\[
    \mathbb E_{\rho}[q_m^{(r)}q_m^{(r)\top}]
    =\frac1{|\mathcal I_m|}
      \sum_{t\in\mathcal I_m}
      g_t(\widetilde y_t^{(r)})g_t(\widetilde y_t^{(r)})^\top.
\]
Taking expectation over resampled targets and applying the score identity gives
the average token Fisher. We use $M=30{,}000$ and $R=4$. Resampling from
$p_0$ and using clean carrier contexts prevents the teacher signal from
contaminating the curvature estimate.

\section{Related Work}
\label{app:related-work}

\paragraph{Subliminal learning.}
Knowledge distillation conventionally transfers a teacher's predictive
behavior by matching its outputs \citep{hinton2015distilling}. Subliminal
learning shows that model-generated data can additionally transmit behavioral
traits that are not semantically expressed in those data
\citep{cloud2025subliminal}. Subsequent work has studied when this transfer
occurs across architectures, carrier tasks, and intervention parameterizations
\citep{schrodi2025towards,blank2026subliminal}. These studies establish the
generality of hidden trait transfer, whereas we focus on the optimization
geometry that determines how accurately a known trait direction can be
recovered.

\paragraph{Steering-vector accounts.}
Recent work connects subliminal transfer to the distillation of latent
steering directions \citep{blank2026subliminal} and introduces subliminal
steering as a controlled way to encode stronger, directly measurable hidden
signals \citep{morgulis2026subliminal}. We adopt this controlled setting because
it exposes both the teacher direction and the recovered student direction. Our
analysis complements these accounts by showing that the carrier task first
transforms the teacher direction through its Fisher geometry and that iterative
optimization acts as an increasingly strong inverse of that transformation.

\paragraph{Data scale and signal localization.}
Empirically, increasing the number of independently generated distillation
examples makes latent teacher traits more recoverable
\citep{dong2026scaling}. Related evidence emphasizes that independent carrier
diversity is more useful than repeatedly training on a fixed dataset and links
transfer to token positions at which teacher predictions diverge
\citep{feng2026everyrank}. These observations explain where trait-dependent
signals can appear, but do not characterize why recovery continues to depend
on large independent datasets after the mean learning direction has stabilized.
Our noisy-inverse account separates these roles: carrier geometry determines
the systematic transformation of the trait, while independent hard targets
control the residual discretization error that is amplified as optimization
inverts that transformation.

\paragraph{Inverse problems and iterative regularization.}
The locally linearized recovery dynamics are a finite-dimensional instance of
Landweber iteration \citep{landweber1951iteration}: the iteration count acts as
a regularization parameter, and each Fisher eigendirection is governed by a
spectral filter. This perspective is classical in inverse problems
\citep{tikhonov1977solutions,engl1996regularization,hansen1998rank}, including
the use of Tikhonov-style damping and early stopping to suppress unstable
low-curvature modes. Statistical inverse-problem theory further characterizes
the corresponding bias--variance tradeoff when the observation error is
random and decreases with sample size
\citep{bissantz2007convergence,cavalier2008nonparametric}. Our plots of trait
energy across the Fisher spectrum are also related to spectral source and
discrete Picard conditions, which describe how the target coefficients are
distributed relative to an ill-posed operator's spectrum
\citep{hansen1998rank}.

\paragraph{Spectral filtering and early stopping in learning.}
Gradient descent has likewise been studied as a spectral regularization method
in learning theory \citep{yao2007early,bauer2007regularization,logerfo2008spectral}.
This literature establishes consistency and stopping rules for iterative
regularization \citep{raskutti2014early}, relates gradient-flow paths to ridge
regularization \citep{ali2019continuous}, and analyzes mode-dependent learning
and overtraining in high-dimensional models \citep{advani2020high}. Our
contribution is therefore not the general observation that gradient descent
implements a spectral filter. We identify the carrier Fisher as the relevant
forward operator for subliminal steering, derive the hard-label forcing term
and its sample-size dependence, and test the resulting mode-wise and
data-scaling predictions in an auditable parameter-recovery setting.

\paragraph{Fisher geometry and distributional supervision.}
Fisher preconditioning is central to natural-gradient methods
\citep{amari1998natural}; importantly, the model Fisher used here is distinct
from the observed-label outer product often called the empirical Fisher
\citep{kunstner2019limitations}. Prior statistical accounts of knowledge
distillation show that replacing sampled hard labels with teacher
probabilities can reduce the variance of the student objective
\citep{menon2021statistical}. In our setting, this variance reduction has a
specific inverse-problem consequence: distributional supervision retains the
carrier-transformed signal while removing the conditional token-sampling noise
that would otherwise be amplified along low-curvature directions.

\section{Hyperparameters}
\label{app:hyperparameters}

Table~\ref{tab:model-data-hparams} reports the model and data configuration.
All pretrained parameters are frozen; only one zero-initialized residual vector
is optimized and shared across the listed layers. Unless stated otherwise, we
use seed 42, BF16 precision, teacher scale $\alpha=1$, and temperature 1.

\begin{table*}[h]
\centering
\small
\setlength{\tabcolsep}{5pt}
\renewcommand{\arraystretch}{1.08}
\begin{tabular}{lcc}
\hline
Configuration & Qwen2.5-7B-Instruct & Gemma-2-9B-IT \\
\hline
Hidden width / transformer layers & $3584$ / $28$ & $3584$ / $42$ \\
Intervened layers & $[2,26)$ & $[2,40)$ \\
Teacher norm bound & $0.5$ & $1.5$ \\
Traits & \multicolumn{2}{c}{cat, dog, eagle, lion, raven} \\
Teacher training examples & \multicolumn{2}{c}{$50$} \\
Teacher optimizer / learning rate & Adam / $10^{-3}$ & Adam / $10^{-3}$ \\
Teacher optimization steps / batch size & $200$ / $10$ & $200$ / $10$ \\
Carrier temperature / top-$k$ & \multicolumn{2}{c}{$1.0$ / full vocabulary (no truncation)} \\
Recovery objectives & \multicolumn{2}{c}{hard NLL and soft KL} \\
Recovery optimizer / learning rate & SGD / $10^{-2}$ & SGD / $0.5$ \\
\hline
\end{tabular}
\caption{Model, intervention, teacher-extraction, carrier-generation, and
student-recovery hyperparameters. The same recovery optimizer and learning rate
are used for hard and soft supervision.}
\label{tab:model-data-hparams}
\end{table*}

\section{Limitations and future work}
\label{app:discussion-limitations}

Our analysis is local. It linearizes the model around the unsteered state and
treats the carrier Fisher matrix $F$ as approximately constant along the
bounded recovery trajectory. This approximation captures the observed
curvature-dependent ordering of recovery, but changes in the model Jacobian and
Fisher geometry can account for deviations from the fixed-$F$ dynamics. 

We use subliminal steering as an experimental platform because it turns hidden
trait transfer into an auditable parameter-recovery problem: both the implanted
teacher direction and the recovered student direction are known, so recovery
can be measured directly rather than inferred only from behavior. Representing
the teacher trait by a single low-dimensional steering vector also makes an
explicit Fisher analysis computationally feasible. This control does not cover
the full range of subliminal learning. Traits induced by system prompts and
students trained through LoRA or full-parameter fine-tuning may involve larger,
higher-dimensional, and more nonlinear changes than the bounded steering
trajectories studied here. Whether the same noise-dominated inverse mechanism
accounts for their data requirements therefore remains to be tested across
broader teacher constructions, student parameterizations, and carrier tasks;
we leave those further validations for future work.

\subsection*{AI use statement}
Generative AI tools were used to assist with code implementation and debugging,
experimental analysis, figure and table preparation, and language editing and
organization of the manuscript. The authors reviewed and verified all
AI-assisted outputs, independently validated the reported experimental results,
and take full responsibility for the scientific claims and final content.

\subsection*{Reproducibility statement}
Code is available at
\url{https://github.com/luoyuchenmlcv/subliminal-data}.

\end{document}